\documentclass{article}
\usepackage{ijcai25}

\usepackage{times}
\usepackage{soul}
\usepackage{url}
\usepackage[hidelinks]{hyperref}
\usepackage[utf8]{inputenc}
\usepackage[small]{caption}
\usepackage{graphicx}
\usepackage{amsmath}
\usepackage{amsthm}
\usepackage{booktabs}
\usepackage{algorithm}
\usepackage{algorithmic}
\usepackage[switch]{lineno}
\usepackage{tikz}

\usepackage{threeparttable}
\usepackage{xspace}
\usepackage{kotex}
\usepackage{multirow}
\usepackage{tablefootnote}
\usepackage{multicol}
\usepackage{amssymb}

\usepackage{mdwlist}

\newcommand{\method}{SPRINT\xspace}
\newcommand{\methodfull}{\underline{S}ublayer \underline{P}\underline{R}uning w\underline{I}th Late\underline{N}cy and \underline{T}unability Information}
\newcommand*{\QED}{\hfill\ensuremath{\square}}%
\newtheorem{problem}{Problem}

\newcommand{\hF}{\widehat{F}}

\newcommand{\vx}{\boldsymbol{x}}

\newcommand{\mW}{\boldsymbol{W}}
\newcommand{\mhW}{\widehat{\boldsymbol{W}}}

\newcommand{\mX}{\boldsymbol{X}}
\newcommand{\mhX}{\widehat{\boldsymbol{X}}}

\newcommand{\mZ}{\boldsymbol{Z}}
\newcommand{\mhZ}{\widehat{\boldsymbol{Z}}}

\title{
    Accurate Sublayer Pruning for Large Language Models \\
    by Exploiting Latency and Tunability Information\
}

\author{
Seungcheol Park$^1$ \and
Sojin Lee$^1$ \and
Jongjin Kim$^1$ \and
Jinsik Lee$^2$ \and
Hyunjik Jo$^2$ \And
U Kang$^1$ 
\\
\affiliations
$^1$Seoul National University\\
$^2$LG AI Research\\
\emails
\{ant6si, lsjlsj5846, j2kim99, ukang\}@snu.ac.kr$^1$,
\{jinsik.lee,hyunjik.jo\}@lgresearch.ai$^2$
}

\begin{document}

\maketitle

\begin{abstract}
How can we accelerate large language models (LLMs) without sacrificing accuracy?
%
%
The slow inference speed of LLMs hinders us to benefit from their remarkable performance in diverse applications.
This is mainly because numerous sublayers are stacked together in LLMs.
Sublayer pruning compresses and expedites LLMs via removing unnecessary sublayers.
However, existing sublayer pruning algorithms are limited in accuracy since they naively select sublayers to prune, overlooking the different characteristics of each sublayer.

In this paper, we propose \method (\methodfull), an accurate sublayer pruning method for LLMs.
\method accurately selects a target sublayer to prune by considering 1) the amount of latency reduction after pruning and 2) the tunability of sublayers.
\method iteratively prunes redundant sublayers and swiftly tunes the parameters of remaining sublayers.
Experiments show that \method achieves the best accuracy-speedup trade-off, 
exhibiting up to 23.88\%p higher accuracy on zero-shot commonsense reasoning benchmarks compared to existing pruning algorithms.

\end{abstract}

\section{Introduction}
\textit{
How can we accelerate large language models (LLMs) without sacrificing accuracy?
}
%
Recent LLMs have shown impressive performance across various tasks such as translation, code completion, and personal assistant~\cite{brown2020language,zhang2022opt,gemini,chowdhery2023palm,llama,llama2,exaone}.
%
The vast number of parameters in LLMs enables the remarkable capabilities, but slows down the inference speed of LLMs, limiting their practical deployments.
Hence, accelerating LLMs is essential to fully leverage their benefits.

Pruning compresses and expedites neural network models via removing unnecessary parameters~\cite{park2024comprehensive,auber}.
LLMs consist of multi-head attention (MHA) and multi-layer perceptron (MLP) sublayers stacked alternatingly.
Sublayer pruning identifies unimportant sublayers, and removes them~\cite{sleb,shortgpt,blockpruner}.
Note that sublayers are sequentially calculated unlike smaller units such as attention heads or neurons which are processed in parallel~\cite{llmpruner,slicegpt}.
Thus, sublayer pruning algorithms display better accuracy-speedup trade-off than finer-grained ones since removing parallelizable computations prevents GPUs from fully utilizing their computational capability.
%

\begin{figure*}[t]
    \centering
    \includegraphics[width=\textwidth]{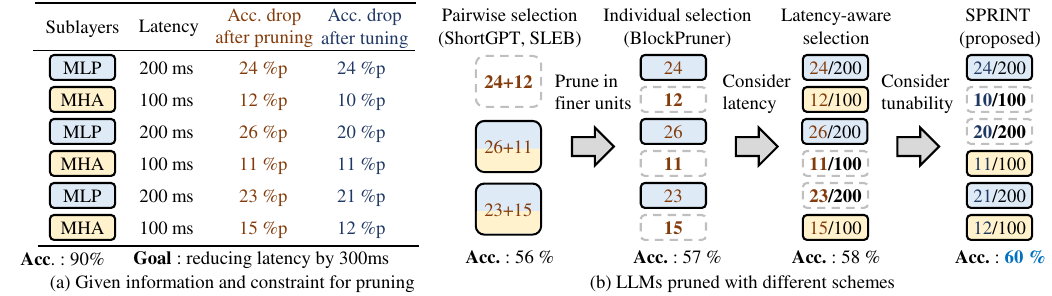}
    \caption{
        (a) Comparison of different schemes in pruning for large language models, given the characteristics of each sublayer.
        Each scheme prunes the least important sublayers judged by its own importance measure.
        (b)
        Among all schemes, \method achieves the best accuracy under the same latency constraint by considering both latency and tunability.
    }
    \vspace{-3mm}
    \label{fig:crown}
\end{figure*}

The core objective of sublayer pruning is to accurately identify the sublayers to prune.
However, existing sublayer pruning algorithms face challenges in preserving the accuracy of a pruned LLM since they fail to consider the characteristics of each sublayer.
Figure~\ref{fig:crown} illustrates the results of different schemes to prune sublayers of an LLM.
%
All the schemes prune less important sublayers to reduce the latency of the model by 300ms.
Figure~\ref{fig:crown}(a) summarizes the attributes of each sublayer, and (b) shows how each scheme evaluates the importance of each sublayer.
Pairwise selection algorithms~\cite{sleb,shortgpt} prune MHA and MLP sublayers in pairs, to reduce the number of importance evaluations for identifying the targets.
Those algorithms face challenges in preserving the accuracy of the pruned LLM since MHA sublayers induce less accuracy drop than MLP sublayers do when pruned.
Individual selection algorithms~\cite{blockpruner}, which evaluate and prune each sublayer separately, still fail to achieve the best accuracy for the following two reasons.
First, they overlook the latency difference between MHA and MLP sublayers.
They prioritize MHA sublayers for removal due to their seemingly lower impact on accuracy.
However, removing a single MLP sublayer provides equivalent latency reduction to removing two MHA sublayers, and they do not consider this effect for accelerating LLMs.
%
Second, they disregard the fact that the accuracy drop caused by pruning is changed after tuning.
They select sublayers solely based on the damage before tuning, failing to find sublayers that cause lower damage after tuning.
%
To overcome these limitations, a pruning algorithm needs to consider both the latency and tunability of each sublayer.

We propose \method (\methodfull), an accurate sublayer pruning method for LLMs.
\method preserves the accuracy of the pruned models via accurately selecting a sublayer to remove.
As shown in Figure~\ref{fig:crown}(b), \method considers the amount of latency reduction after pruning and accuracy drop after tuning.
Moreover, we minimize the cost of \method by 1) activation checkpointing to mitigate the repetitive computations, and 2) fast candidate selection to reduce the number of time-intensive tuning.
We verify the effectiveness of \method with extensive experiments.
\method accelerates Llama-2 and Llama-3 models, achieving up to 23.88\%p higher accuracy on zero-shot commonsense reasoning benchmarks than baselines, and shows the best accuracy-speedup trade-off.

We summarize our main contributions as follows:
\begin{itemize*}
    \item \textbf{Algorithm.}
    We propose \method, an accurate sublayer pruning method for LLMs.
    \method accurately identifies less important sublayers in a model by the following four effective techniques: 1) latency-aware importance scoring, 2) tunability-aware sensitivity evaluation, 3) activation checkpointing, and 4) fast candidate selection.

    \item \textbf{Experiments.}
    We demonstrate that \method achieves the state-of-the-art performance on commonsense reasoning benchmarks.
    \method accelerates Llama-2 and Llama-3 models, achieving up to 23.88\%p higher accuracy on zero-shot commonsense reasoning benchmarks, showing the most favorable accuracy-speedup trade-off.

    \item \textbf{Analysis.}
    We analyze the pruning patterns of the models pruned by \method.
    We derive the findings that MLP and lower sublayers in an LLM serve as a critical component of LLM's capabilities, while MHA and upper sublayers contribute less to accuracy.

\end{itemize*}

The rest of this paper is organized as follows.
We first define LLM acceleration problem and provide backgrounds.
We then propose \method, our pruning method.
After presenting experimental results, we conclude.
Our source code is available at \url{https://github.com/snudm-starlab/SPRINT}. 
%

\section{Preliminary}

\begin{figure*}[t]
    \centering
    \includegraphics[width=\textwidth]{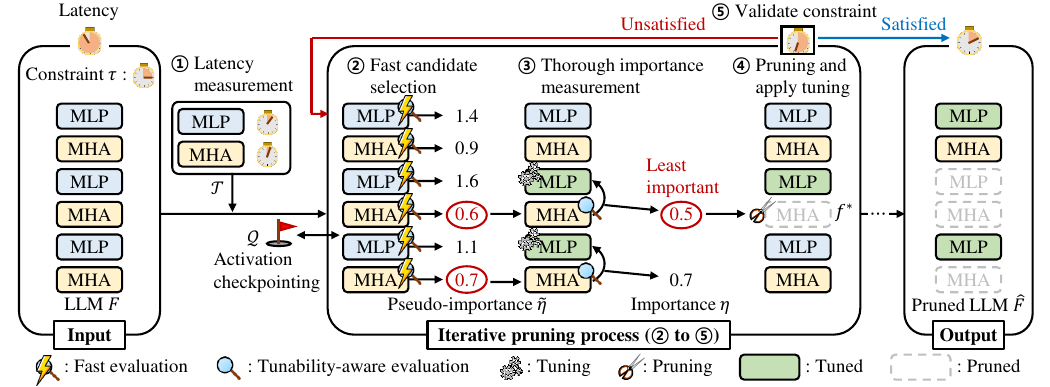}
    \caption{
        An illustration of the overall process of \method.
        Given a pretrained LLM and a latency constraint, \method iteratively identifies and prunes the least important sublayer until the pruned model satisfies the latency constraint.
        \method accurately selects the sublayer to prune by considering the latency and tunability information of sublayers.
        }
    \vspace{-3mm}
    \label{fig:method}
\end{figure*}

\subsection{Problem Definition}
\begin{problem}[LLM Acceleration Problem]
    Given a pretrained large language model $F$, a sample dataset $\mathcal{D}$, and a latency upper bound $\tau$,
    the problem is to find an accurate model $\hF$ whose latency does not exceed $\tau$.
    \QED
\end{problem}

\subsection{Transformer Architecture}
Recent LLMs~\cite{llama2,llama3} have a Transformer-based architecture~\cite{Transformer} which consists of multi-head attention (MHA) and multi-layer perceptron (MLP) sublayers stacked alternatingly.
%
%
A Transformer model $F$ with $S$ sublayers refines an input sequence vector $\vx$ as in Equation~\eqref{eq:Transformer}.
\begin{equation}
    \label{eq:Transformer}
        F(\vx) = \mathcal{G}
        \left(
        \left(
        \circ_{s=1}^{S} (
            f^{(s)} + I
        )
        \right)
        \Bigl(
            \mathcal{E}(\vx)
        \Bigr)
        \right)
\end{equation}
%
%
$\mathcal{G}$ is a generator module,
$\mathcal{E}$ is an embedding look-up table,
$I$ represents a residual connection,
and $\circ$ is the composition of functions.
$f^{(s)}$ denotes the $s$th sublayer function which is either a
self-attention network in an MHA sublayer when $s$ is an odd number,
or a feed-forward network in an MLP sublayer when $s$ is an even number.
We decompose $f^{(s)}$ into output projection $\mW^{(s)}$ and remainder $h^{(s)}(\cdot)$ as in Equation~\eqref{eq:decomp}.
\begin{equation}
	\label{eq:decomp}
    f^{(s)}(\mX^{(s)}) = \mW^{(s)}h^{(s)}(\mX^{(s)})
\end{equation}
$\mX^{(s)}$ is an input matrix for the $s$th sublayer, where $\mX^{(1)} = \mathcal{E}(\vx)$.
Given an intermediate activation $\mZ^{(s)}=h^{(s)}(\mX^{(s)})$, $f^{(s)}$ is a linear transformation function with regard to $\mZ^{(s)}$.


\subsection{Sublayer Pruning}
Sublayer pruning~\cite{sleb,shortgpt,blockpruner} accelerates LLMs via pruning unnecessary sublayers in them.
Sublayer pruning algorithms measure an importance score $\eta$ for each sublayer, and eliminate those with the lowest scores.
The importance scoring leverages sensitivity $\zeta$ which represents the performance difference of a model before and after the pruning. 
It is crucial to accurately score the importance of sublayers since the accuracy of the model heavily depends on which sublayers are pruned.
\subsection{Fast In-compression Tuning}
\label{subsec:incomp_tuning}
Pruning causes the accuracy loss by repeatedly removing parameters from a model.
To mitigate the accuracy degradation, tuning the pruned model is essential so that the inference results are similar to those of the uncompressed model~\cite{kprune,park2024comprehensive}.
The objective of the tuning is to align the output of the $s$th sublayer in the pruned model with that in the unpruned model, as described in Equation~\eqref{eq:3lstsq}.
\begin{eqnarray}
    \label{eq:3lstsq}
    \arg\min_{
        \mhW^{(s)}
        }
        \|
            (\mhX^{(s)}+\mhW^{(s)}\mhZ^{(s)})-\mX^{(s+1)}
        \| _F^2,
\end{eqnarray}
where $\mhX^{(s)}$, $\mhW^{(s)}$, and $\mhZ^{(s)}$ are the input, output projection, and intermediate representation of the $s$th sublayer in the pruned model, respectively.
$\mhX^{(s+1)}=\mhX^{(s)}+\mhW^{(s)}\mhZ^{(s)}$ is the output of $s$th sublayer in the pruned model, while $\mX^{(s+1)}$ is the output of the $s$th sublayer in the unpruned model.

Note that solving Equation~\eqref{eq:3lstsq} does not require time-consuming stochastic gradient descents as in previous works~\cite{lora,qalora}. 
Instead, the equation is efficiently solved incorporating PyTorch's solver (torch.linalg.lstsq) once $\mhX^{(s)}$, $\mhZ^{(s)}$, and $\mX^{(s+1)}$ are computed.
Thus, fast in-compression tuning is computationally affordable for iterative pruning algorithms.

\section{Proposed Method}

\begin{figure*}[t]
    \centering
    \includegraphics[width=0.95\textwidth]{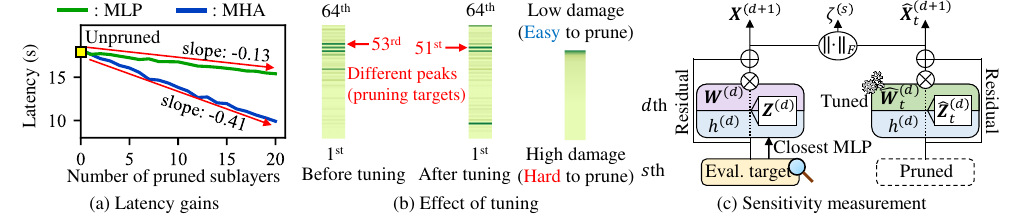}
    \vspace{-2mm}
    \caption{
        (a) Latency change after pruning MHA and MLP sublayers in Llama-3 8B model.
        MHA sublayers impact more than MLP sublayers after pruning.
        (b) The rankings of sublayers according to the amount of accuracy loss caused by pruning, before and after tuning.
        Tuning impacts the rankings, changing pruning targets.
        (c) An illustration of tunability-aware sensitivity measurement process.
        \method performs fast in-compression tuning
        on the closest upper MLP sublayer while measuring sensitivities.
    }
    \label{fig:456}
    \vspace{-3mm}
\end{figure*}

\subsection{Overview}
We address the following challenges to prune sublayers in LLMs minimizing the loss of accuracy.
\begin{itemize*}
    \item [C1.] \textbf{Latency Difference of Sublayers.}
    Existing works ignore the latency difference of MHA and MLP sublayers.
    How can we compare the importance of sublayers with different latencies to effectively accelerate LLMs?
    \item [C2.] \textbf{Ignoring the Impact of Tuning.}
    Existing works ignore the impact of tuning when selecting sublayers to prune, resulting in misselection.
    How can we incorporate the impact of tuning to accurately select sublayers to prune?
    \item [C3.] \textbf{Expensive Computational Cost.}
    Sublayer pruning is computationally expensive since it repeatedly measures the importance scores of all sublayers at each iteration.
    How can we enhance the efficiency of sublayer pruning?
\end{itemize*}
We propose \method to address these challenges.
The main ideas of \method are as follows.
\begin{itemize*}
    \item [I1.] \textbf{Latency-aware Importance Scoring.}
    We consider the amount of latency reduction after pruning each sublayer to precisely identify unimportant sublayers.
    \item [I2.] \textbf{Tunability-aware Sensitivity Evaluation.}
    We accurately select sublayers to prune by measuring their sensitivity after tuning.
    \item [I3.] \textbf{Avoiding Unnecessary Computations.}
    We propose activation checkpointing and fast candidate selection to avoid the unnecessary computations.
\end{itemize*}

\renewcommand{\algorithmicrequire}{\textbf{Input:}}
\renewcommand{\algorithmicensure}{\textbf{Output:}}
\renewcommand{\algorithmicindent}{0.7em}


\begin{algorithm}[t]
	\caption{Overall process of \method}
	\begin{algorithmic}[1]
		\REQUIRE An LLM $F$,
            a calibration dataset $\mathcal{D}$,
            a latency constraint $\tau$,
            number $\alpha$ of checkpoints, and
            number $\beta$ of candidates
		\ENSURE A pruned LLM $\hF$
        \STATE Initialize $\hF$ as $F$
        \STATE Initialize a dictionary $\mathcal{Q}$ of $\alpha$ checkpoints
        \STATE Measure latencies $\mathcal{T}=\{t^{(MHA)}$,\;$t^{(MLP)}\}$
        \WHILE {$\text{latency}(\hF)>\tau$}
            \STATE $\mathcal{C}, \mathcal{Q}\leftarrow
            \text{fast\_candidate\_selection}(\hF, \mathcal{D}, \mathcal{T},\beta, \mathcal{Q})$
            \item[] $\quad \rhd$ Select a set $\mathcal{C}$ of candidates (Section~\ref{sec:eff})
            \STATE $f^*\leftarrow \text{tunability\_aware\_target\_selection}(\mathcal{C}, \hF, \mathcal{D}, \mathcal{T}, \mathcal{Q})$
            \item[] $\quad\rhd$ Find the least important sublayer $f^*$ (Section~\ref{sec:tun})
            \STATE Remove $f^*$ from $\hF$ and apply tuning
        \ENDWHILE
		\RETURN $\hF$
	\end{algorithmic}
	\label{alg:method}
\end{algorithm}

Algorithm~\ref{alg:method} and Figure~\ref{fig:method} show the overall process of \method.
Given a pretrained LLM $F$ and a latency constraint $\tau$, \method returns the pruned LLM $\hF$ satisfying the latency constraint.
\method initializes $\alpha$ activation checkpoints $\mathcal{Q}$ to store reusable activations (line 2, details in Section~\ref{sec:eff}).
Then, \method measures the amounts $\mathcal{T}$ of latency reduction resulting from the removal of sublayers (line 3) for latency-aware importance scoring (details in Section~\ref{sec:lat}).
\method repeats the iterative process of 1) scoring the importance of sublayers and 2) removing the least important one (lines 4-8) until the latency constraint is met.
%
For each iteration, \method first selects $\beta$ candidate sublayers $\mathcal{C}$ to prune by scoring the pseudo-importance of each sublayer (line 5, see Section~\ref{sec:eff}), and updates the checkpoints $\mathcal{Q}$.
\method then scores the importance of each candidate in $\mathcal{C}$ with tunability-aware sensitivity evaluation (line 6, details in Section~\ref{sec:tun}).
Based on the importance scores, \method prunes the least important sublayer $f^*$ and tunes the remaining model (line 7).
%
\method returns the pruned model $\hF$ (line 9) which satisfies the constraint.
\subsection{Latency-aware Importance Scoring}
\label{sec:lat}

\paragraph{Observation.}
How can we identify the most appropriate sublayer to prune for accelerating LLMs with minimal accuracy loss?
Existing works~\cite{sleb,blockpruner,shortgpt} measure the importance of sublayers without considering the latency difference of sublayers.
Figure~\ref{fig:456}(a) shows the latencies of Llama-3 8B models after pruning different numbers of MHA and MLP sublayers.
As shown in the figure,
pruning an MHA sublayer yields over three times greater latency reduction than pruning an MLP sublayer;
thus, it is beneficial to prune an MHA sublayer instead of an MLP sublayer if they cause the same damage.
Therefore, it is essential to consider latency when selecting sublayers to prune.

\paragraph{Our solution.}
We incorporate the latencies of sublayers into our importance scoring process and assign lower importance scores for the sublayers with higher latencies to promote pruning the high-latency sublayers.
The importance $\eta^{(s)}$ of the $s$th sublayer is defined as follows:
\begin{equation}
    \eta^{(s)} = {\zeta^{(s)}}/{t^{(s)}},
\end{equation}
where $\zeta^{(s)}$ is the sensitivity of the $s$th sublayer which approximates the amount of accuracy degradation after pruning it.
$t^{(s)}$ is the amount of reduced latency through pruning the $s$th sublayer.
Hence, $\eta^{(s)}$ reflects the cost-effectiveness of the $s$th sublayer in contributing to accuracy.
As shown in Figure~\ref{fig:456}(a), sublayers of the same type offer almost the same degree of latency reduction; $t^{(s)}$ is either $t^{(MHA)}$ or $t^{(MLP)}$ depending on the sublayer's type.
\method measures $t^{(MHA)}$ and $t^{(MLP)}$ by comparing the latencies of unpruned and partially pruned models before starting its iterative pruning process.


\begin{figure*}
    \centering
    \includegraphics[width=0.95\textwidth]{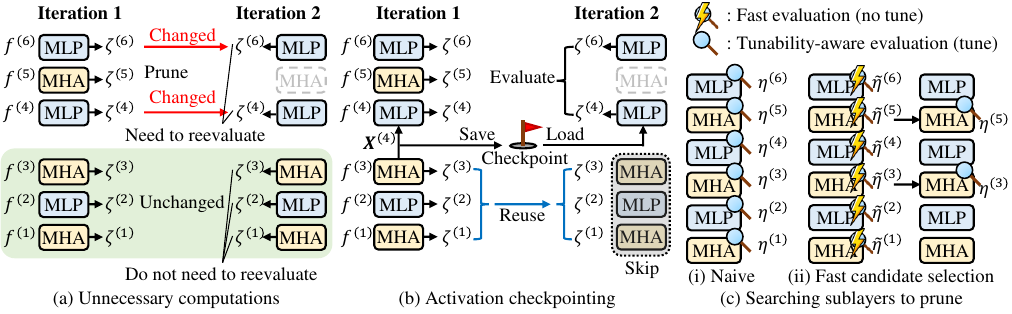}
    \vspace{-2mm}
    \caption{
        (a) Unnecessary computations in iterative sublayer pruning.
        The sensitivity of a sublayer is unchanged if its closest upper MLP sublayer is below the pruned sublayer.
        (b) An illustration of activation checkpointing.
        \method avoids unnecessary computations by caching sensitivities and activations in the previous iterations.
        (c) Comparison between the naive approach and the fast candidate selection to find the sublayer to prune.
        The fast candidate selection reduces the number of layers to tune.
    }
    \label{fig:789}
    \vspace{-3mm}
\end{figure*}

\subsection{Tunability-aware Sensitivity Evaluation}
\label{sec:tun}
\paragraph{Observation.}
How can we accurately estimate the sensitivities of sublayers?
Sublayer pruning algorithms measure sensitivities to approximately estimate the accuracy loss after pruning each sublayer.
The lost accuracy is recovered via tuning, and each sublayer has a different capability for recovering.
However, existing works~\cite{shortgpt,sleb,blockpruner} ignore the effect of tuning when estimating the sensitivities.
Figure~\ref{fig:456}(b) compares the ranking in accuracy degradation after pruning each sublayer before and after tuning.
As shown in the figure, the index of the peak sublayer that evokes the lowest damage is changed after tuning.
This indicates that pruning sublayers with the lowest sensitivity without considering the effect of tuning removes useful sublayers that exhibit low accuracy degradation after tuning.
Therefore, sublayer pruning algorithms must account for the effect of tuning by prioritizing the removal of sublayers that result in minimal accuracy drop after tuning.

\paragraph{Our solution.}
\method compares the activations of the original model and the pruned model after tuning to measure the sensitivities of sublayers.
\method exploits the fast in-compression tuning~\cite{kprune} in Section~\ref{subsec:incomp_tuning} to efficiently incorporate tunability information into the sublayer selection process.
Figure~\ref{fig:456}(c) shows the sensitivity measurement process of \method for the $s$th sublayer which we call as the evaluation target.
\method finds the closest MLP sublayer ($d$th sublayer in the Figure~\ref{fig:456}(c)) above the evaluation target.
After that, \method measures the sensitivity $\zeta^{(s)}$ of the $s$th sublayer by computing the normalized distance between outputs $\mX^{(d+1)}$ and $\mhX^{(d+1)}_t$ of the MLP sublayer before and after pruning, respectively, as in Equation~\eqref{eq:zeta}.
We exploit $\mhX^{(d+1)}_t$ obtained by fast in-compression tuning to take the tunability into account.
\begin{equation}
    \label{eq:zeta}
    \zeta^{(s)} = ||\mX^{(d+1)} - \mhX^{(d+1)}_t||_F / ||\mX^{(d+1)}||_F,
\end{equation}
Note that we use outputs of only MLP sublayers,
since an MLP sublayer has three times more number of parameters than that of an MHA sublayer,
and thus has a stronger tuning capability than MHA.
For fast in-compression tuning, \method finds $\mhW^{(d)}_t$ that minimizes the distance between the output $\mhX^{(d+1)}= (\mhX^{(d)}+\mhW^{(d)}\mhZ^{(d)})$ after pruning and the output $\mX^{(d+1)}$ before pruning, as in Equation~\eqref{eq:3lstsq}.
$\mhW^{(d)}$ and $\mhZ^{(d)}$ are the weight of the out projection and the intermediate activation of the $d$th sublayer after pruning, respectively.
Note that each row of $\mhW^{(d)}$ forms an independent subproblem and we tune the weights in only $c$\% of rows to avoid overfitting.
    We select the rows with abundant outliers, which represent larger activations than others, to maximize the impact of tuning with the given percentage of rows.
    We exploit the sum of the outlier-aware weight-wise scores~\cite{wanda,owl} of weights in each row for selection.
%
We save the tuned weights of each sublayer during the importance scoring process and apply the tuned weights corresponding to the pruning of the least important sublayer.

\subsection{Avoiding Unnecessary Computations}
\label{sec:eff}

\paragraph{Observation 1.}
How can we minimize the computation for measuring the sensitivities of sublayers?
%
Pruning a sublayer does not affect the sensitivity of other sublayers whose closest upper MLP sublayers are located below the pruned one.
For instance, as shown in Figure~\ref{fig:789}(a), $\zeta^{(1)}$ to $\zeta^{(3)}$ do not need to be reevaluated if $f^{(5)}$ is pruned in the previous iteration. 
However, naively computing the sensitivities of all sublayers at each iteration entails redundant computation for sensitivities which are already obtained in the previous iteration.
%

\paragraph{Our solution 1 (Activation Checkpointing).}
We propose activation checkpointing to prevent unnecessary recomputations.
Before starting the iterations, \method places checkpoints between sublayers.
At each iteration, \method stores the activations at the checkpoints.
\method reuses the sensitivities from the previous iteration for each sublayer $f$ 
whose closest upper MLP sublayer is beneath the pruned sublayer, since the sensitivity of $f$ is not changed after pruning.
Then, \method updates the sensitivities of the remaining sublayers using the stored activation. 
For example,
assume we pruned $f^{(5)}$ at iteration 1, as shown in Figure~\ref{fig:789}(b).
Note that
\method reuses $\zeta^{(1)}$ to $\zeta^{(3)}$ from the iteration 1 in the iteration 2
since they are not changed.
Then, \method loads the stored activation $\mX^{(4)}$ and starts updating the sensitivities from $f^{(4)}$.
The checkpoints are uniformly placed, and the number $\alpha$ of checkpoints controls the trade-off between the memory usage and the time consumption during pruning. 


\paragraph{Observation 2.}
How can we minimize the cost of sensitivity measurement?
\method performs an in-compression tuning to evaluate the sensitivity of each sublayer.
Naively computing the sensitivities of all sublayers as illustrated in Figure~\ref{fig:789}(c-i) leads to excessive number of tunings at each iteration,
making the sensitivity measurement too expensive.

\paragraph{Our solution 2 (Fast Candidate Selection).}
We propose fast candidate selection to selectively measure the sensitivities, minimizing the number of tuning in the sensitivity measurement.
Instead of evaluating tunability-aware sensitivities for all sublayers,
\method first selects the candidates of the least sensitive sublayers swiftly without tuning.
Then, \method measures the tunability-aware sensitivity only for the candidate sublayers.
For example, in Figure~\ref{fig:789}(c-ii), \method first finds two candidates without tuning and then selects the sublayer to prune with the tunability-aware evaluation, reducing the number of tunings from 6 to 2.
This process can be viewed as reducing the computational costs by approximately finding the sublayer to prune.

To instantly select the candidate sublayers, 
\method measures the pseudo-importance of each sublayer.
The pseudo-sensitivity $\tilde{\zeta}^{(s)}$ of $s$th sublayer is the normalized distance between outputs $\mX^{(d+1)}$ and $\mhX^{(d+1)}$ of the MLP sublayer before and after pruning, respectively.
Note that $\mhX^{(d+1)}$ is obtained without tuning, unlike $\mhX_t^{(d+1)}$ in Equation~\eqref{eq:zeta}.
The pseudo-importance $\tilde{\eta}^{(s)}$ of the $s$th sublayer is $\tilde{\zeta}^{(s)}/t^{(s)}$, where $t^{(s)}$ is the latency of the sublayer.
\method selects $\beta$ sublayers with the least pseudo-importance scores, where
$\beta$ is a hyperparameter representing the number of candidates.
A higher $\beta$ leads to the more accurate search for the sublayer to prune while requiring a higher computational cost.

\begin{figure*}[t]
    \centering
    \includegraphics[width=0.96\textwidth]{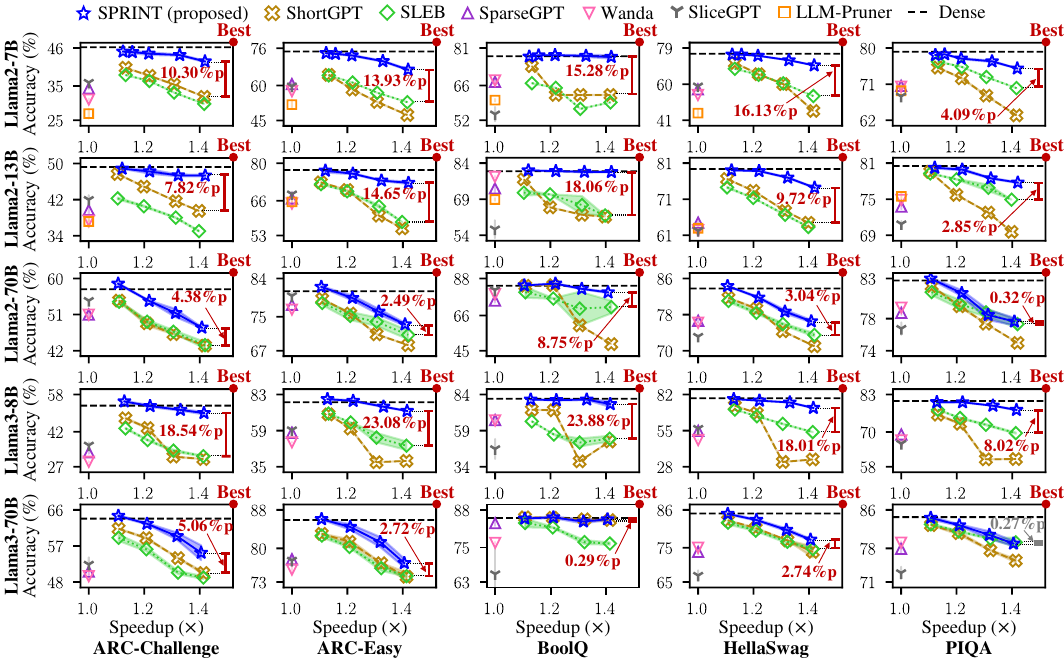}
    \vspace{-2mm}
    \caption{
        Accuracy-speedup trade-off curves of \method and competitors.
        \method shows the best trade-off among all the methods.
    } \vspace{-4mm}
    \label{fig:perf}
\end{figure*}

\section{Experiments}
We perform experiments to answer the following questions.
\begin{itemize*}
    \item [Q1.] \textbf{Accuracy.} How accurate is \method compared to baselines with the similar acceleration level?
    \item [Q2.] \textbf{Pruning Efficiency.}
    How fast does \method prune LLMs compared to baselines?
    \item [Q3.] \textbf{Ablation Study.} Does each main idea of \method contribute to the performance?
    \item [Q4.] \textbf{Pruning Pattern Analysis.} Which sublayers are important to maintain the accuracy of LLMs?
\end{itemize*}

\subsection{Experimental Setup}
\paragraph{Setup.}
We use Llama-2~\cite{llama2} and Llama-3~\cite{llama3} model families as pruning targets.
We randomly sample 128 token sequences of length 2048 from Wikitext2~\cite{wikitext} dataset for sensitivity measurement and tuning.
We use NVIDIA A100 80GB GPU for all experiments.
We report zero-shot reasoning accuracies on ARC-Challenge, ARC-Easy~\cite{arc}, BoolQ~\cite{boolq}, HellaSwag~\cite{hellaswag}, and PIQA~\cite{piqa} benchmarks.
We measure the latencies to generate 512 tokens from 1024 input tokens~\cite{Qserve} and report speedups of pruned models.
%

\paragraph{Baselines.}
We compare \method with three sublayer pruning algorithms:
ShortGPT~\cite{shortgpt}, SLEB~\cite{sleb}, and BlockPruner~\cite{blockpruner}.
ShortGPT and SLEB utilize the pairwise selection scheme while BlockPruner selects pruning targets individually as illustrated in Figure~\ref{fig:crown}.
We also include four fine-grained pruning algorithms as baselines for comprehensive analysis:
SparseGPT~\cite{sparsegpt}, Wanda~\cite{wanda}, SliceGPT~\cite{slicegpt}, and LLM-Pruner~\cite{llmpruner}.
These algorithms prune LLMs in parallelizable units smaller than sublayers such as channels or attention heads,
resulting in insufficiently accelerated models.

%

\begin{figure*}[t]
    \centering
    \includegraphics[width=0.96\textwidth]{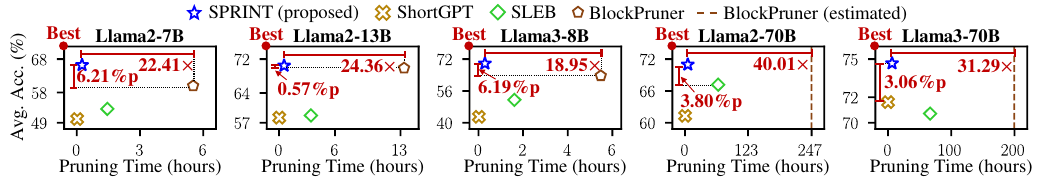}
    \vspace{-1mm}
    \caption{
            Pruning time and accuracy trade-off of \method and competitors under 1.4$\times$ acceleration.
            \method is closest to the ``Best" point with the highest accuracy and short pruning time.
    }
    \label{fig:cost}
    \vspace{-4mm}
\end{figure*}

\paragraph{Hyperparameters.}
We use random seeds from 0 to 2 and report the average values.
We set the checkpointing hyperparameter $\alpha$ to 8 and the candidate hyperparameter $\beta$ to 5 for all models.

\subsection{Accuracy}
Figure~\ref{fig:perf} shows the accuracies of \method and baselines across various inference speedup settings.
As shown in the figure, \method achieves the best trade-off curve among all the methods, and significantly outperforms the competitors under 40\% speedup setting with the maximum accuracy gap of 23.88\%p.
It is notable that \method achieves the best performance among all the algorithms on challenging multiple-choice benchmarks, such as ARC-Challenge, ARC-Easy, and HellaSwag, across all models of various sizes.

\subsection{Pruning Efficiency}
Figure \ref{fig:cost} visualizes compression time and average accuracy on five commonsense reasoning tasks of the pruned models generated by \method and competitors with 40\% speedup.
We estimate the pruning time of BlockPruner for 70B models since it takes more than a week to prune them (see Supplementary Material for details).
Brown dashed lines in Figure~\ref{fig:cost} denote the estimated pruning time.

Note that \method offers the best accuracy-pruning time trade-off across all models.
\method prunes the LLM up to 40.01 times faster 
and achieves up to 6.21\%p higher accuracy than BlockPruner.
\method further outperforms other methods with similar pruning times by larger margins.

%

\begin{table}[t]
    \centering 
    \begin{threeparttable}
        \resizebox{\columnwidth}{!}{
            \begin{tabular}{l|rrr|r}
                \toprule
                Method & \method-$l$ & \method-$t$ & \method-$e$ & \method \\
                \midrule
                Acc. (\%) & 66.96 & 67.62 & 69.82 & 69.82 \\
                Time (s) & 1070.29 & 212.00 & 2147.28 & 1052.21 \\
                \bottomrule
            \end{tabular}
            }
        \end{threeparttable}
        \vspace{-1mm}
    \caption{
    Performance of \method and its variants for accelerating Llama-3 8B by 1.4$\times$.
    See Section~\ref{sec:ablation} for details.
    }\vspace{-3mm}
    \label{tab:ablation}
\end{table}

\subsection{Ablation Study}
\label{sec:ablation}
To prove the effectiveness of each main idea, we compare \method with its three variants: \method-$l$, \method-$t$, and \method-$e$.
\method-$l$ is \method without latency-aware importance scoring; it prunes the sublayer with the lowest sensitivity without considering the latency.
\method-$t$ is \method without tunability-aware sensitivity evaluation, measuring the sensitivity of each sublayer by directly comparing its output before and after pruning.
\method-$e$ is \method without activation checkpointing and fast candidate selection,
calculating the importance scores of all sublayers without storing activations for each step of iterative pruning process.

Table~\ref{tab:ablation} summarizes the performance of \method and its variants when accelerating the Llama-3 8B by 40\%.
\method and \method-$e$ outperform other variants in terms of accuracy, proving that both
latency-aware importance scoring and tunability-aware sensitivity evaluation contribute to the accuracy.
\method is twice faster than \method-$e$, confirming the effectiveness of activation checkpointing and fast candidate selection for the pruning cost.
In summary, all three main ideas of \method contributes to the performance.

\subsection{Pruning Pattern Analysis}
\label{sec:pattern}
\begin{figure}[t]
    \centering
    \includegraphics[width=1\columnwidth]{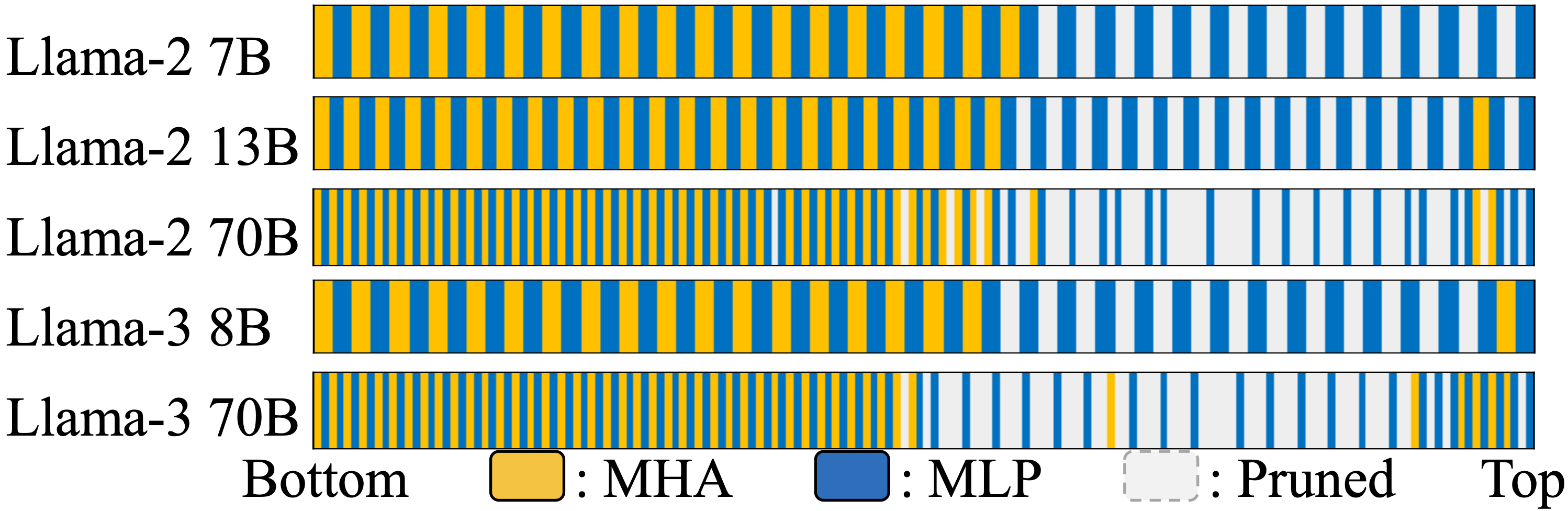}
    \caption{
        Pruning patterns of \method on Llama models (best viewed in color).
        \method primarily prunes MHA sublayers located between the middle and upper parts of the model.
    }
    \label{fig:pattern}
    \vspace{-3mm}
\end{figure}

Figure~\ref{fig:pattern} depicts the pruning patterns of \method on Llama models.
Orange, blue, and gray squares represent MHA, MLP, and pruned sublayers, respectively.
We observe two main patterns related to the type and position of sublayers.
First, \method prunes more MHA sublayers than MLP sublayers in general;
MLP sublayers are pruned only in 70B models.
Second, \method prunes sublayers located mainly in the upper-middle parts of models.
These two patterns show that MLP sublayers and sublayers located near the bottom significantly contribute to the capability of LLMs.
Moreover, as more extensive pruning occurs in the upper layers as observed by the second pattern,
sensitivities of lower layers often do not need to be updated.
Thus, our proposed activation checkpointing aligns well with the LLM's characteristics.

\section{Related Work}


%
%

We review techniques for accelerating LLMs: quantization, knowledge distillation, and dynamic inference.
Quantization~\cite{sensimix,optq,flexround,awq,omni,synq,zsqsurvey} reduces the bit-width of weights and activations in them;
it accelerates computation by leveraging hardwares designed for low-bit operations.
Quantization is compatible with pruning;
unifying both methods achieves greater acceleration~\cite{sparsegpt}.

Knowledge distillation (KD)~\cite{distillm,kegnet,peakd,ddk,jhkim21,pet,falcon} improves the accuracy of compressed models by transferring knowledge from uncompressed models.
KD is also compatible with pruning, by effectively compensating for the error induced by pruning.
%

Dynamic inference~\cite{calm,dynamic3,mod,lazyllm} accelerates LLMs via dynamically adjusting the amount of computations based on the input.
Dynamic inference exhibits minimal accuracy degradation since it determines the amount of computations according to the importance of inputs.
However, they have a significant drawback in that their efficiency is diminished when multiple inputs requiring different computations are fed~\cite{sleb}.
In contrast, sublayer pruning consistently preserves efficiency regardless of the number of inputs.

\vspace{0.5mm}
\section{Conclusion}
\vspace{0.5mm}

We propose \method, an accurate sublayer pruning method for accelerating LLMs.
\method addresses the inaccurate sublayer selection problem of existing sublayer pruning methods
by factoring in latency and tunability information.
We propose activation checkpointing and fast candidate selection techniques to shorten the running time of \method.
We demonstrate that \method achieves the best accuracy-speedup trade-off when pruning Llama-2 and Llama-3 models.

\newpage

\section*{Contribution Statement}
Seungcheol Park, Sojin Lee, and Jongjin Kim contributed equally to this work as co-first authors.
Jinsik Lee and Hyunjik Jo provided helpful discussion for developing our method.
U Kang supervised the project and carefully reviewed the manuscript.

\section*{Acknowledgements}
This work was supported by Institute of Information \& communications Technology Planning \& Evaluation (IITP) grant funded by the Korea government (MSIT)
[No.RS-2020-II200894, Flexible and Efficient Model Compression Method for Various Applications and Environments],
[No.RS-2021-II211343, Artificial Intelligence Graduate School Program (Seoul National University)],
and [No.RS-2021-II212068, Artificial Intelligence Innovation Hub (Artificial Intelligence Institute, Seoul National University)].
This work was supported by Youlchon Foundation.
The Institute of Engineering Research at Seoul National University provided research facilities for this work.
The ICT at Seoul National University provides research facilities for this study.
U Kang is the corresponding author.
This work was improved by the helpful input and collaboration of researchers from LG AI Research.

\bibliographystyle{named}
\bibliography{paper}

\newpage

\appendix

\section{Symbols}
We summarize the definitions of the symbols in Table~\ref{tab:sym}.
\begin{table}[h]
    \centering
    \begin{threeparttable}
        \begin{tabular}{c|l}
        \toprule
        Symbol & Definition \\
        \midrule
        $F$ & Transformer model \\
        $\mathcal{G}$ & generator module \\
        $\mathcal{E}$ & embedding look-up table \\
        $\vx$ & input sequence of tokens \\
        \midrule
        $S$ & number of sublayers \\
        $f^{(s)}$ & $s$th sublayer \\
        $\mX$ & input of a sublayer\\
        $\mZ$ & intermediate activation of a sublayer\\
        $\mW$ & output projection of a sublayer\\
        $\widehat{\mX}$ & input in the pruned model\\
        $\widehat{\mZ}$ & intermediate activation in the pruned model\\
        $\widehat{\mW}$ & output projection in the pruned model\\
        \midrule
        $t$ & a latency \\
        $\zeta$ & sensitivity \\
        $\eta$ & importance score \\
        $\tilde{\zeta}$ & pseudo sensitivity without tuning\\
        $\tilde{\eta}$ & pseudo importance score without tuning\\
        \midrule
        $c$ & channel-selection hyperparameter \\
        $\alpha$ & checkpointing hyperparameter \\
        $\beta$ & candidate hyperparameter \\
        \midrule
        $\circ$ & composition of multiple functions\\
        \bottomrule
        \end{tabular}
    \end{threeparttable}
    \caption{Symbols and definitions.}
    \label{tab:sym}
\end{table}

\section{Detailed Experimental Setup}
\subsection{Evaluation Protocol}
We run each experiment thrice with the random seeds in $[0, 1, 2]$.
We then report the average and the standard deviation.
We summarize detailed evaluation protocols for accuracy and acceleration as follows.

\subsubsection{Accuracy.}
For accuracy evaluation, we use five commonsense reasoning benchmarks spanning diverse domains to comprehensively assess the capability of models.
We summarize the statistics of each dataset in Table~\ref{tab:data}.
We utilize evaluation source codes in language model evaluation harness~\cite{eval-harness}.
\begin{itemize}
    \item \textbf{ARC-Challenge and ARC-Easy}~\cite{arc} are datasets composed of grade-school level multiple-choice science questions.
    The challenge set contains only questions that neither a retrieval-based algorithm nor a word co-occurrence algorithm correctly answers.
    The easy set contains the remaining questions.
    Each problem consists of a scientific question and four choices.
    \item \textbf{BoolQ}~\cite{boolq} is a question-answering dataset whose questions are generated in unprompted and unconstrained settings.
    It provides a passage and a question; the model should answer the question with either yes or no.
    \item \textbf{HellaSwag}~\cite{hellaswag} presents a short passage and four phrases for each problem.
    A model needs to choose the most appropriate phrase to complete the given passage.
    \item \textbf{PIQA}~\cite{piqa} tests the physical commonsense knowledge of a model.
    It gives two possible solutions for a given question to a model, and lets the model choose the most appropriate solution.
\end{itemize}

\begin{table}[h]
    \centering
    \begin{threeparttable}
        \begin{tabular}{lrr}
        \toprule
        Dataset & Instance & Choices \\
        \midrule
        ARC-Challenge & 1,172 & Multiple (4) \\
        ARC-Easy & 2,376 & Multiple (4) \\
        BoolQ & 3,270 & Binary (2) \\
        HellaSwag & 10,003 & Multiple (4) \\
        PIQA & 3,000 & Binary (2) \\
        \bottomrule
        \end{tabular}
    \end{threeparttable}
    \caption{Statistics of datasets.}
    \label{tab:data}
\end{table}

\subsubsection{Acceleration.}

We report acceleration performance based on the latency required to generate 512 tokens, given a context length of 1024.
We set the size of the batch to 1 when we measure the latencies of models.
We use a single GPU to evaluate Llama-2 7B, 13B and Llama-3 8B.
We use two GPUs to evaluate Llama-2 70B and Llama-3 70B.

For fine-grained pruning methods, we use the settings with the highest accuracies among those reported in SLEB~\cite{sleb}.
Experimental results of SLEB demonstrate that fine-grained pruning methods underperform SLEB
even for a larger batch size of 64.

\subsection{Device Settings}
We use a machine with
a CPU (Intel Xeon Gold 6338 2g (32C/64T)), 8 64GB RAMs,
and 4 GPUs (NVIDIA A100 80GB).
The machine is operated by Ubuntu 22.04.3 LTS.
%


\subsection{Implementation Detail of \method}

\subsubsection{Multi-GPU Protocol.}
We summarize the GPU usage of \method during compression in Table~\ref{tab:gpu}.
We utilize two GPUs to compress and evaluate 70B models.
We store 40 layers in DRAM, and load only 40 layers at a time into VRAM to implement the multi-GPU protocol.
In this way, \method secures the VRAM space for the in-compression tuning by keeping unnecessary sublayers on CPU.

\begin{table}[h]
    \centering
    \begin{threeparttable}
        \begin{tabular}{l|ccc|cc}
        \toprule
           \multirow{2}{*}{Method} & \multicolumn{3}{c|}{Llama-2} & \multicolumn{2}{c}{Llama-3} \\
            & 7B & 13B & 70B & 8B & 70B \\
         \midrule
            SparseGPT & 1 & 1 & 3 & 1 & 3 \\
            Wanda & 1 & 1 & 2 & 1 & 2 \\
            SliceGPT & 1 & 1 & 2 & 1 & 2 \\
            LLM-Pruner & 1 & 1 & 2 & 1 & 2 \\
            ShortGPT & 1 & 1 & 2 & 1 & 2 \\
            SLEB & 1 & 1 & 2 & 1 & 2 \\
            BlockPruner & 1 & 1 & 2 & 1 & 2 \\
            \method & 1 & 1 & 2 & 1 & 2 \\
        \bottomrule
        \end{tabular}
    \end{threeparttable}
    \caption{GPU usage of pruning algorithms.}
    \label{tab:gpu}
\end{table}


\subsubsection{Channel-selection Hyperparameter.}

We search the optimal channel-selection hyperparameter $c$ in the range of $[25, 50, 75, 100]$.
Note that pruning a sublayer induces stronger damage respect to the number of parameters in the output projection for smaller models such as Llama-2 7B, 13B, and Llama-3 8B.
Hence, we set $c$ to $100$ for small models.
For 70B models, we evaluate \method on Llama-3 70B when the model is accelerated by 40\% with varying $c$.
Table~\ref{tab:gamma} shows the results.
    The pruned model shows the highest accuracy when $c$ is $75$, thus we set $c$ as $75$ for 70B models.
%

\begin{table}[h]
    \centering
    \begin{threeparttable}
        \begin{tabular}{c|cccc}
        \toprule
        $c$ & 25 & 50 & 75 & 100 \\
        \midrule
        Avg. acc. & 69.06 & 72.87 & \textbf{74.52} & 73.55 \\
        \bottomrule
        \end{tabular}
    \end{threeparttable}
    \caption{Average accuracy of \method over different channel-selection hyperparameter $c$.}
    \label{tab:gamma}
\end{table}


\subsubsection{Candidate Hyperparameter.}

We evaluate \method on Llama-3 8B when the model is accelerated by 40\%, adjusting the candidate hyperparameter $\beta$ to values of $[1, 3, 5, 7, 9]$.
We report the average accuracy of each setting after pruning 7 sublayers and 13 sublayers in Table~\ref{tab:m}.
Note that the accuracy remains constant when $\beta$ is $5$ or greater.
This demonstrates that five candidates are sufficient to find the sublayer with the least importance score.
Hence, we set $\beta$ to  $5$ for all experiments.

\begin{table}[h]
    \centering
    \begin{threeparttable}
        \begin{tabular}{c|ccccc}
        \toprule
        $\beta$ & 1 & 3 & 5 & 7 & 9 \\
        \midrule
        Step 7 & 74.14 & 74.19 & 74.19 & 74.19 & 74.19  \\
        Step 13 & 70.71 & 70.66 & 70.71 & 70.71 & 70.71  \\
        \bottomrule
        \end{tabular}
    \end{threeparttable}
    \caption{Hyperparameter search for $\beta$.}
    \label{tab:m}
\end{table}

\subsection{Implementation Detail of Competitors}

\subsubsection{Implementation.}
We use the official implementations publicly available online for
SparseGPT\footnotemark[1]~\cite{sparsegpt},
Wanda\footnote{\url{https://github.com/locuslab/wanda}}~\cite{wanda},
LLM-Pruner\footnote{\url{https://github.com/horseee/LLM-Pruner}}~\cite{llmpruner},
SliceGPT\footnote{\url{https://github.com/microsoft/TransformerCompression}}~\cite{slicegpt},
and SLEB\footnote{\url{https://github.com/jiwonsong-dev/SLEB}}~\cite{sleb}.
We modify the sensitivity measurement of SLEB to implement ShortGPT~\cite{shortgpt}.
We change the pruning granularity of SLEB to implement BlockPruner~\cite{blockpruner}.
We follow the hyperparameter settings of the original paper for every baseline.
We summarize the GPU usage of each method during compression in Table~\ref{tab:gpu}.

\subsubsection{Execution Time of BlockPruner.}
BlockPruner spends more than a week to prune Llama-2 70B and Llama-3 70B due to its high time complexity.
Thus, we report only the estimated pruning time of BlockPruner in our paper.
BlockPruner scores the importance of each sublayer during each iteration by evaluating the perplexity on a sample dataset after removing the sublayer.
Hence, the execution time of BlockPruner is predicted by estimating the latency required for perplexity measurement.

Suppose a model retains $n_1$ MHA sublayers and $n_2$ MLP sublayers, where the computation time of an MHA sublayer and an MLP sublayer during the perplexity measurement are $t^{(MHA)}$ and $t^{(MLP)}$, respectively.
Then, the time requirement to measure the perplexity of the model is as follows:
\begin{equation}
    n_1t^{(MHA)}+n_2t^{(MLP)} \label{eq:model}
\end{equation}

To assess the sensitivities of all sublayers, BlockPruner needs to measure the sensitivity of MHA sublayer for $n_1$ times and the sensitivity of MLP sublayer for $n_2$ times.
BlockPruner measures the sensitivity of each sublayer by evaluating the perplexity of the remaining model when the sublayer is pruned.
Hence, the time $T^{(MHA)}$ for measuring an MHA sublayer's sensitivity is $(n_1-1)t^{(MHA)}+n_2t^{(MLP)}$,
and the time $T^{(MLP)}$ for measuring an MLP sublayer's sensitivity is $n_1t^{(MHA)}+(n_2-1)t^{(MLP)}$.
Thus, the execution time $t$ for a single iteration of BlockPruner is as follows:
\begin{eqnarray}
    t =& n_1T^{(MHA)}+n_2T^{(MLP)} \nonumber\\
    =& \quad n_1\left((n_1-1)t^{(MHA)}+n_2t^{(MLP)}\right) \nonumber\\
    &+\; n_2\left(n_1t^{(MHA)}+(n_2-1)t^{(MLP)}\right) \nonumber\\
    =& (n_1+n_2-1)(n_1t^{(MHA)}+n_2t^{(MLP)}). \label{eq:bp}
\end{eqnarray}
%
%

To obtain $t^{(MHA)}$ and $t^{(MLP)}$, we set up equations for the inference time using Equation~\eqref{eq:bp}. 
We collect two data points from the first two iterations of BlockPruner as shown in Table~\ref{tab:bplat}.
Then, we solve the equations to calculate $t^{(MHA)}$ and $t^{(MLP)}$.

\begin{table}[h]
    \centering
    \begin{threeparttable}
        \begin{tabular}{c|c|ccc}
        \toprule
        Model & Step & $n_1$ & $n_2$ & Time (s) \\
        \midrule
        \multirow{2}{*}{Llama-2 70B}
        & 1 & 80 & 80 & 30764.47 \\
        & 2 & 79 & 80 & 30455.34 \\
        \midrule
        \multirow{2}{*}{Llama-3 70B}
        & 1 & 80 & 80 & 31357.19 \\
        & 2 & 79 & 80 & 31064.53 \\
        \bottomrule
        \end{tabular}
    \end{threeparttable}
    \caption{Time requirement for each step of BlockPruner.}
    \label{tab:bplat}
\end{table}

We estimate the lower bound of BlockPruner's pruning time by assuming that BlockPruner prunes only MLP sublayers after the first two iterations until the latency constraint is met. 
This enables achieving the latency constraint by pruning fewer sublayers than pruning MHA sublayers.

\section{Detailed Experimental Results}
We summarize our detailed experimental results in Tables~\ref{tab:perf_2-7B} to~\ref{tab:perf_3-70B}.
The values in parentheses denote the standard deviations.
LLM-Pruner cannot prune Llama-2 70B and Llama-3 8B, 70B since it does not support group query attention.
We do not report the results of BlockPruner for 70B models due to its excessive pruning cost.

\begin{table*}[h]
    \centering
    \begin{threeparttable}
    \begin{tabular}{c|c|r|ccccc|c}
        \toprule
        {Method} & {Speedup} & Pruning time (s) &
        ARC-C & ARC-E & BoolQ & HellaSwag & PIQA & Avg.\\
         \midrule
        Unpruned       &                   1.0x        & - & 46.25 & 74.58 & 77.74 & 76.01 & 79.11 & 70.74  \\
        \midrule
        SparseGPT   &                 1.0x          & 540.85 & 34.30 & 60.62 & 67.46 & 57.24 & 70.60 & 58.05 ($\pm$0.08) \\
        \midrule
        Wanda       &                 1.0x          & 65.42 & 31.03 & 57.52 & 67.98 & 54.27 & 70.40 & 56.24 ($\pm$0.05) \\
        \midrule
        LLM-Pruner  &                 1.0x          & 72.70 & 26.93 & 51.84 & 60.07 & 44.76 & 69.37 & 50.59 ($\pm$1.29) \\
        \midrule
        SliceGPT    &                   1.0x        & 437.19 & 35.98 & 59.72 & 54.46 & 58.93 & 67.97 & 55.41 ($\pm$0.73) \\
        \midrule
        \multirow{4}{*}{ShortGPT} &      1.1x       & 78.05 & 40.27 & 64.59 & 73.75 & 71.15 & 75.14 & 64.98 ($\pm$0.01) \\
                    &                    1.2x       & 71.44 & 37.94 & 58.28 & 62.17 & 65.27 & 72.47 & 59.23 ($\pm$0.03) \\
                    &                   1.3x        & 71.80 & 35.41 & 52.96 & 62.17 & 59.99 & 68.15 & 55.74 ($\pm$0.00) \\
                    &                   1.4x        & 71.25 & 31.83 & 47.35 & 62.17 & 45.92 & 63.22 & 50.10 ($\pm$0.00) \\
        \midrule
        \multirow{4}{*}{SLEB}   &       1.1x        & 2550.08 & 38.17 & 64.35 & 66.90 & 68.08 & 76.62 & 62.83 ($\pm$0.02) \\
                     &                   1.2x       & 3583.54 & 36.43 & 61.28 & 65.04 & 64.92 & 75.32 & 60.60 ($\pm$0.02) \\
                    &                   1.3x        & 4471.81 & 33.08 & 56.90 & 56.60 & 60.42 & 72.78 & 55.96 ($\pm$0.02) \\
                    &                   1.4x        & 5225.36 & 29.78 & 52.86 & 59.20 & 53.73 & 70.02 & 53.12 ($\pm$0.00) \\
        \midrule
        \multirow{4}{*}{BlockPruner} &    1.1x      & 7201.90 & 45.56 & 74.21 & 78.22 & 75.52 & 78.24 & 70.35 ($\pm$0.18) \\
                 &                        1.2x      & 11028.77 & 44.34 & 71.58 & 74.32 & 73.33 & 78.15 & 68.34 ($\pm$0.62) \\
                 &                       1.3x       & 16584.32 & 40.19 & 64.37 & 69.17 & 69.67 & 77.31 & 64.14 ($\pm$0.34) \\
                 &                       1.4x       & 19872.34 & 36.97 & 60.28 & 62.99 & 64.99 & 74.85 & 60.02 ($\pm$0.10) \\
        \midrule
        \multirow{4}{*}{\method (ours)} &  1.1x     & 415.83 & 44.82 & 73.58 & 78.07 & 75.50 & 78.45 & 70.09 ($\pm$0.06) \\
                &                          1.2x     & 669.76 & 44.40 & 72.92 & 78.15 & 74.95 & 77.38 & 69.56 ($\pm$0.00) \\
                          &                1.3x     & 779.18 & 43.97 & 70.44 & 77.87 & 72.59 & 76.70 & 68.31 ($\pm$0.06) \\
                          &                1.4x     & 886.57 & 42.12 & 66.79 & 77.45 & 69.87 & 74.92 & 66.23 ($\pm$0.13) \\
        \bottomrule
    \end{tabular}
    \end{threeparttable}
    \caption{Performance of \method and baselines on Llama-2 7B.}
    \label{tab:perf_2-7B}
\end{table*}

\begin{table*}[h]
    \centering
    \begin{threeparttable}

        \begin{tabular}{c|c|r|ccccc|c}
            \toprule
            {Method} & {Speedup} & Pruning time (s) &
            ARC-C & ARC-E & BoolQ & HellaSwag & PIQA & Avg.\\
             \midrule
             Unpruned       &                   1.0x        & - & 49.23 & 77.48 & 80.58 & 79.39 & 80.52 & 73.44 \\
            \midrule
            SparseGPT   &                 1.0x          & 923.78 & 39.70 & 66.86 & 73.53 & 64.56 & 73.78 & 63.69 ($\pm$0.60) \\
            \midrule
            Wanda       &                 1.0x          & 110.39 & 37.34 & 64.97 & 78.32 & 62.45 & 74.90 & 63.60 ($\pm$0.17) \\
            \midrule
            LLM-Pruner  &                 1.0x          & 127.40 & 37.14 & 65.57 & 68.88 & 62.96 & 75.46 & 62.00 ($\pm$0.66) \\
            \midrule
            SliceGPT    &                   1.0x        & 725.00 & 41.98 & 68.48 & 56.53 & 62.30 & 70.87 & 60.03 ($\pm$0.54) \\
            \midrule
            \multirow{4}{*}{ShortGPT} &      1.1x       & 101.59 & 47.67 & 72.88 & 77.25 & 76.61 & 79.54 & 70.79 ($\pm$0.01) \\
                        &                    1.2x       & 94.59 & 44.88 & 70.03 & 65.62 & 73.34 & 75.73 & 65.92 ($\pm$0.00) \\
                        &                   1.3x        & 94.29 & 41.64 & 60.45 & 62.54 & 67.78 & 72.78 & 61.04 ($\pm$0.00) \\
                        &                   1.4x        & 93.79 & 39.51 & 55.68 & 61.74 & 64.48 & 69.59 & 58.20 ($\pm$0.00) \\
            \midrule
            \multirow{4}{*}{SLEB}   &       1.1x        & 4976.16 & 42.26 & 72.22 & 71.69 & 74.28 & 79.29 & 67.95 ($\pm$0.25) \\
                         &                   1.2x       & 8328.47 & 40.53 & 69.56 & 71.01 & 71.35 & 78.31 & 66.15 ($\pm$0.00) \\
                        &                   1.3x        & 10500.17 & 38.00 & 63.95 & 66.57 & 66.69 & 76.90 & 62.42 ($\pm$0.99) \\
                        &                   1.4x        & 12317.69 & 35.01 & 58.07 & 62.23 & 63.32 & 74.92 & 58.71 ($\pm$0.02) \\
            \midrule
            \multirow{4}{*}{BlockPruner} &    1.1x      & 19296.92 & 49.49 & 76.56 & 80.21 & 79.37 & 80.49 & 73.22 ($\pm$0.08) \\
                     &                        1.2x      & 30716.04 & 48.46 & 76.22 & 80.15 & 79.32 & 80.14 & 72.86 ($\pm$0.00) \\
                     &                       1.3x       & 41103.73 & 46.33 & 73.99 & 80.03 & 78.08 & 79.60 & 71.61 ($\pm$0.00) \\
                     &                       1.4x       & 48204.63 & 44.60 & 71.89 & 76.87 & 75.73 & 78.85 & 69.89 ($\pm$0.00) \\
            \midrule
            \multirow{4}{*}{\method (ours)} &  1.1x     & 799.33 & 48.86 & 77.26 & 81.01 & 79.02 & 80.23 & 73.28 ($\pm$0.17) \\
                    &                          1.2x     & 1237.05 & 48.24 & 76.12 & 80.65 & 78.81 & 79.92 & 72.75 ($\pm$0.12) \\
                              &                1.3x     & 1711.71 & 47.35 & 73.56 & 80.37 & 76.99 & 78.47 & 71.35 ($\pm$0.03) \\
                              &                1.4x     & 1979.15 & 47.33 & 72.71 & 80.30 & 74.20 & 77.77 & 70.46 ($\pm$0.10) \\
            \bottomrule
        \end{tabular}
        \end{threeparttable}
    \caption{Performance of \method and baselines on Llama-2 13B.}
    \label{tab:perf_2-13B}
\end{table*}

\begin{table*}[h]
    \centering
    \begin{threeparttable}

        \begin{tabular}{c|c|r|ccccc|c}
            \toprule
            {Method} & {Speedup} & Pruning time (s) &
            ARC-C & ARC-E & BoolQ & HellaSwag & PIQA & Avg.\\
             \midrule
             Unpruned       &                   1.0x        & - & 57.34 & 80.98 & 83.73 & 83.80 & 82.75 & 77.72 ($\pm$3.51) \\
            \midrule
            SparseGPT   &                 1.0x          & 5056.86 & 51.00 & 77.78 & 74.88 & 76.65 & 78.71 & 71.80 ($\pm$0.47) \\
            \midrule
            Wanda       &                 1.0x          & 506.33 & 51.00 & 76.68 & 78.60 & 76.22 & 79.36 & 72.37 ($\pm$0.20) \\
            \midrule
            SliceGPT    &                   1.0x        & 3550.79 & 54.47 & 79.91 & 81.17 & 73.14 & 76.75 & 73.09 ($\pm$0.19) \\
            \midrule
            \multirow{4}{*}{ShortGPT} &      1.1x       & 400.15 & 54.49 & 79.19 & 83.65 & 81.72 & 82.06 & 76.23 ($\pm$0.02) \\
                        &                    1.2x       & 400.59 & 48.81 & 75.74 & 84.07 & 79.30 & 79.54 & 73.49 ($\pm$0.12) \\
                        &                   1.3x        & 402.60 & 46.13 & 70.86 & 60.10 & 74.32 & 77.37 & 65.76 ($\pm$0.00) \\
                        &                   1.4x        & 400.50 & 43.26 & 68.31 & 48.93 & 71.00 & 74.97 & 61.29 ($\pm$0.00) \\
            \midrule
            \multirow{4}{*}{SLEB}   &       1.1x        & 95869.04 & 54.21 & 78.25 & 79.80 & 81.04 & 81.30 & 74.92 ($\pm$0.73) \\
                        &                   1.2x       & 155204.11 & 49.23 & 75.18 & 76.08 & 78.41 & 79.80 & 71.74 ($\pm$0.23) \\
                        &                   1.3x        & 197089.86 & 46.73 & 73.82 & 70.05 & 75.91 & 78.78 & 69.06 ($\pm$1.86)  \\
                        &                   1.4x        & 232648.35 & 43.37 & 70.72 & 70.99 & 73.49 & 77.31 & 67.18 ($\pm$0.39) \\
            \midrule
            BlockPruner &   1.4x        & 888537.14  & - & -& -& -& -& - \\
            \midrule
            \multirow{4}{*}{\method (ours)}
            &  1.1x     & 10114.22 & 58.70 & 82.04 & 83.39 & 83.85 & 84.33 & 78.38 ($\pm$0.01) \\
            &  1.2x     & 14823.28 & 54.41 & 79.42 & 82.57 & 84.95 & 81.79 & 76.35 ($\pm$0.66) \\
            &  1.3x     & 17109.60 & 51.39 & 76.19 & 79.03 & 81.83 & 78.67 & 73.31 ($\pm$0.48) \\
            &  1.4x     & 19334.55 & 47.75 & 73.11 & 73.21 & 79.74 & 77.64 & 70.97 ($\pm$0.19) \\
            \bottomrule
        \end{tabular}
        \end{threeparttable}
    \caption{Performance of \method and baselines on Llama-2 70B.
        We do not report the accuracy of BlockPruner due to its excessive pruning cost.
    }
    \label{tab:perf_2-70B}
\end{table*}

\begin{table*}[h]
    \centering
    \begin{threeparttable}
        \begin{tabular}{c|c|r|ccccc|c}
            \toprule
            {Method} & {Speedup} & Pruning time (s) &
            ARC-C & ARC-E & BoolQ & HellaSwag & PIQA & Avg.\\
             \midrule
             Unpruned       &                   1.0x        & - & 53.24 & 77.82 & 81.07 & 79.16 & 80.74 & 74.41 \\
            \midrule
            SparseGPT   &                 1.0x          & 649.14 & 33.19 & 57.63 & 66.33 & 54.88 & 69.17 & 56.24 ($\pm$0.51) \\
            \midrule
            Wanda       &                 1.0x          & 70.34 & 29.12 & 51.19 & 66.01 & 47.96 & 67.08 & 52.28 ($\pm$0.15) \\
            \midrule
            SliceGPT    &                   1.0x        & 467.53 & 36.21 & 59.41 & 46.33 & 56.53 & 66.00 & 52.89 ($\pm$1.94) \\
            \midrule
            \multirow{4}{*}{ShortGPT} &      1.1x       & 73.25 & 47.81 & 69.80 & 73.50 & 73.60 & 76.01 & 68.14 ($\pm$0.06) \\
                        &                    1.2x       & 82.94 & 43.63 & 60.38 & 73.06 & 68.06 & 72.89 & 63.60 ($\pm$0.02) \\
                        &                   1.3x        & 83.71 & 31.34 & 38.01 & 37.76 & 31.58 & 60.57 & 39.85 ($\pm$0.03) \\
                        &                   1.4x        & 82.95 & 30.29 & 38.89 & 51.53 & 33.45 & 60.66 & 42.96 ($\pm$0.00) \\
            \midrule
            \multirow{4}{*}{SLEB}   &       1.1x        & 2804.23 & 43.49 & 70.33 & 65.44 & 70.73 & 77.71 & 65.54 ($\pm$0.10) \\
                         &                   1.2x       & 3949.30 & 38.45 & 64.37 & 56.34 & 65.79 & 74.88 & 59.97 ($\pm$0.00) \\
                        &                   1.3x        & 4938.45 & 33.93 & 54.34 & 50.77 & 60.03 & 72.54 & 54.32 ($\pm$0.24) \\
                        &                   1.4x        & 5783.30 & 31.51 & 48.99 & 53.46 & 53.95 & 69.64 & 51.51 ($\pm$0.26) \\
            \midrule
            \multirow{4}{*}{BlockPruner} &    1.1x      & 7524.14 & 54.01 & 79.12 & 81.19 & 78.52 & 81.01 & 74.77 ($\pm$0.00) \\
                     &                        1.2x      & 11586.27 & 53.50 & 78.54 & 77.13 & 76.50 & 80.25 & 73.18 ($\pm$0.26) \\
                     &                       1.3x       & 15302.71 & 50.26 & 76.64 & 73.64 & 75.92 & 79.98 & 71.29 ($\pm$0.00) \\
                     &                       1.4x       & 19713.00 & 40.64 & 65.78 & 67.32 & 67.94 & 76.44 & 63.63 ($\pm$0.38) \\
            \midrule
            \multirow{4}{*}{\method (ours)} &  1.1x     & 494.43 & 55.03 & 79.92 & 80.55 & 78.58 & 80.34 & 74.89 ($\pm$0.03) \\
                    &                          1.2x     & 698.27 & 53.04 & 78.76 & 80.08 & 77.43 & 80.32 & 73.93 ($\pm$0.03) \\
                              &                1.3x     & 895.59 & 51.39 & 74.90 & 80.53 & 76.05 & 79.13 & 72.40 ($\pm$0.06) \\
                              &                1.4x     & 1040.51 & 50.06 & 72.07 & 77.34 & 71.96 & 77.66 & 69.82 ($\pm$0.26) \\
            \bottomrule
        \end{tabular}
        \end{threeparttable}
    \caption{Performance of \method and baselines on Llama-3 8B.}
    \label{tab:perf_3-8B}
\end{table*}

\begin{table*}[h]
    \centering    
    \begin{threeparttable}

        \begin{tabular}{c|c|r|ccccc|c}
            \toprule
            {Method} & {Speedup} & Pruning time (s) &
            ARC-C & ARC-E & BoolQ & HellaSwag & PIQA & Avg.\\
             \midrule
             Unpruned       &                   1.0x        & - & 63.82 & 85.90 & 85.17 & 84.93 & 84.49 & 80.86 \\
            \midrule
            SparseGPT   &                 1.0x          & 4911.39 & 50.60 & 77.75 & 83.42 & 73.80 & 77.89 & 72.69 ($\pm$0.42) \\
            \midrule
            Wanda       &                 1.0x          & 506.63 & 49.49 & 75.58 & 76.58 & 75.02 & 79.18 & 71.17 ($\pm$0.34) \\
            \midrule
            SliceGPT    &                   1.0x        & 3958.11 & 52.39 & 77.50 & 65.78 & 66.91 & 72.91 & 67.10 ($\pm$2.56) \\
            \midrule
            \multirow{4}{*}{ShortGPT} &      1.1x       & 401.82 & 61.29 & 83.02 & 85.41 & 82.96 & 82.83 & 79.10 ($\pm$0.02) \\
                        &                    1.2x       & 401.61 & 58.96 & 81.38 & 85.26 & 80.69 & 81.16 & 77.49 ($\pm$0.01) \\
                        &                   1.3x        & 401.73 & 53.98 & 76.92 & 84.70 & 77.13 & 77.58 & 74.06 ($\pm$0.00) \\
                        &                   1.4x        & 354.15 & 50.26 & 74.12 & 84.46 & 73.79 & 75.46 & 71.62 ($\pm$0.75) \\
            \midrule
            \multirow{4}{*}{SLEB}   &       1.1x        & 98505.34 & 59.04 & 82.66 & 83.47 & 82.30 & 83.01 &  78.10($\pm$0.58) \\
                         &                   1.2x       & 159587.78 & 56.20 & 80.29 & 81.94 & 79.98 & 81.54 &  75.99($\pm$0.15) \\
                        &                   1.3x        & 210620.41 & 50.37 & 75.88 & 76.84 & 76.73 & 80.03 &  71.97($\pm$0.22) \\
                        &                   1.4x        & 239478.45 & 49.12 & 74.23 & 76.34 & 74.62 & 79.27 &  70.72($\pm$0.26) \\
            \midrule
            BlockPruner &   1.4x        & 718513.34  & - & -& -& -& -& - \\
            \midrule
            \multirow{4}{*}{\method (ours)}
            & 1.1x & 11079.35 & 64.51 & 86.07 & 85.02 & 84.83 & 84.35 & 80.95 ($\pm$0.17) \\
            & 1.2x & 17026.39 & 62.54 & 84.46 & 85.42 & 83.06 & 82.81 & 79.66 ($\pm$0.18) \\
            & 1.3x & 20426.70 & 59.44 & 81.33 & 83.98 & 80.20 & 80.8 & 77.15 ($\pm$0.52) \\
            & 1.4x & 22962.68 & 55.32 & 76.95 & 84.75 & 77.36 & 79.00 & 74.67 ($\pm$0.63) \\
            \bottomrule
        \end{tabular}
        \end{threeparttable}
    \caption{Performance of \method and baselines on Llama-3 70B.
        We do not report the accuracy of BlockPruner due to its excessive pruning cost.}
    \label{tab:perf_3-70B}
\end{table*}

\end{document}